# AUTOMATED IDENTIFICATION OF DISASTER NEWS FOR CRISIS MANAGEMENT USING MACHINE LEARNING

Lord Christian Carl H. Regacho, Ai Matsushita,

Angie M. Ceniza-Canillo, PhD

Department of Computer, Information Sciences and Mathematics,
University of San Carlos, Cebu City, Philippines, 6000

## ABSTRACT

*A lot of news sources picked up on Typhoon Rai (also known locally as Typhoon Odette), along with fake news outlets. The study honed in on the issue, to create a model that can identify between legitimate and illegitimate news articles. With this in mind, we chose the following machine learning algorithms in our development: Logistic Regression, Random Forest and Multinomial Naive Bayes. Bag of Words, TF-IDF and Lemmatization were implemented in the Model. Gathering 160 datasets from legitimate and illegitimate sources, the machine learning was trained and tested. By combining all the machine learning techniques, the Combined BOW model was able to reach an accuracy of 91.07%, precision of 88.33%, recall of 94.64%, and F1 score of 91.38% and Combined TF-IDF model was able to reach an accuracy of 91.18%, precision of 86.89%, recall of 94.64%, and F1 score of 90.60%.*

## 1. INTRODUCTION

The Philippines consists of hundreds of islands, and it is vulnerable to many types of disasters such as coastal flooding, coastal ecosystem biodiversity loss, cyclones and storm surge. Small island communities are the most exposed to the impacts of natural disasters. Typhoons cost the country about 2% of the yearly GDP on average, and another 2% is needed for the recovery activities. Better prevention policies and adaptation actions could ease the impact of such climate-related risks [1].

The typhoon Rai (local name 'Odette') caused a significant amount of damage not only to the island of Cebu, but to the Philippines as a whole. Prominent media outlets reported on the devastation the typhoon caused. Along with this, a significant number of phoney news outlets released illegitimate news articles about the typhoon Odette.

These fraudulent new outlets desired to capitalise on the anxiety and weary people by publishing fake news articles with exaggerated new titles. One major issue with this is the prevalence of fake disaster news articles, in the hopes of scaring and misinforming random viewers as a form of scare tactics to gain attention. These fraudulent new articles cause significant harm to the people. The researchers decided to pursue this study in identifying legitimate from illegitimate news articles, starting with the recent typhoon Rai. As researchers we do not expect to solve the issue of misinformation but in what we do we promote integrity and honesty when it comes to informing the masses about a specific topic.

## 2. Related Literature

This chapter is composed of the studies that discuss the existing literature and systems from an in-depth perspective  from foreign and local developments related to the thesis topic.

### 2.1. Disaster Media

Numerous natural disasters on an unprecedented level over the past two decades have been covered by mainstream news outlets and various other media and this has caught the attention of various academics as well as the public [2]. The option to analyse the new media website as a location for disaster management and recovery is supplemented by studies that show the strength of media to convey and trivialise cultural assumptions [3].

There is a high amount of disaster data gathered from the Philippines as it is always at risk for natural disasters occuring. Due to this, it is hard to manually sort out all of the information from the most useful ones and the most important ones [4]. The local news website coverage of disaster events is an invaluable cultural resource for affected people and communities in identifying the 'appropriate way of acting to and recovering from a devastating disaster [5].

### 2.3. Natural Language Processing

Natural Language Processing (NLP) are techniques that allow computers to analyse and obtain meaning from human conversations. Although the field of NLP has existed since the 1950s, the development of technology and similar techniques throughout the years has made Natural Language Processing applications simpler to implement, with some functions outperforming human capabilities [6]. Computational Techniques using Natural Language Processing (NLP) automatically create a synopsis on an occurrence of flood given a group of varying news sources. The computational techniques utilise NLTK Python Library and Apache Hadoop to go through and summarise a corpus [7].

An automated natural language processing (NLP) technique looks for, compile, characterise, structure, and collate news coverage descriptions of disaster recovery which are important to grow the appeal of the analysis of text-heavy corpora in furthering the study of pre and post disaster planning [8].

### 2.4. Machine Learning

This being useful by obtaining relevant news from the mainstream news websites and utilising machine learning algorithms, the news is categorised into disaster relevant or irrelevant news. Different algorithms are implemented such as multinomial Naive Bayes, Logistic Regression, SVM, extreme gradient boosting and random forest. The Logistic Regression gave the best results out of all of them [9].

Machine Learning algorithms have progressively been used for tackling scientific and engineering problems. These data driven algorithms have proved immensely useful in producing landslide susceptibility mapping. It has given hopeful results, specifically for mapping landslide prone areas with limited geotechnical data [10]. To differentiate various machine learning algorithms for landslide susceptibility mapping, landslide susceptibility maps are created using conventional (Logistic Regression) and modern techniques (Decision Tree). Results indicate that the spatial paradigm of susceptibility map from logistic regression is continually distributed, on the other hand that the decision tree is short and concise [11].

## 2.5. Text Mining

It is interpreted as the new text mining problem and current general probabilistic technique for solving this issue through the discovery of latent themes from text, creating an evolution graph of themes; and the analysis of life cycles of themes. The evaluation of the aforementioned technique on two varying domains (i.e., news articles and literature) show that the aforementioned can uncover interesting and evolutionary theme patterns [12].

## 2.6. Logistic Regression

In recent years, Logistic regression has been chosen as the analytic technique for the multivariate modelling of categorical dependent variables. However, this procedure is still obscure for many potential users [13].

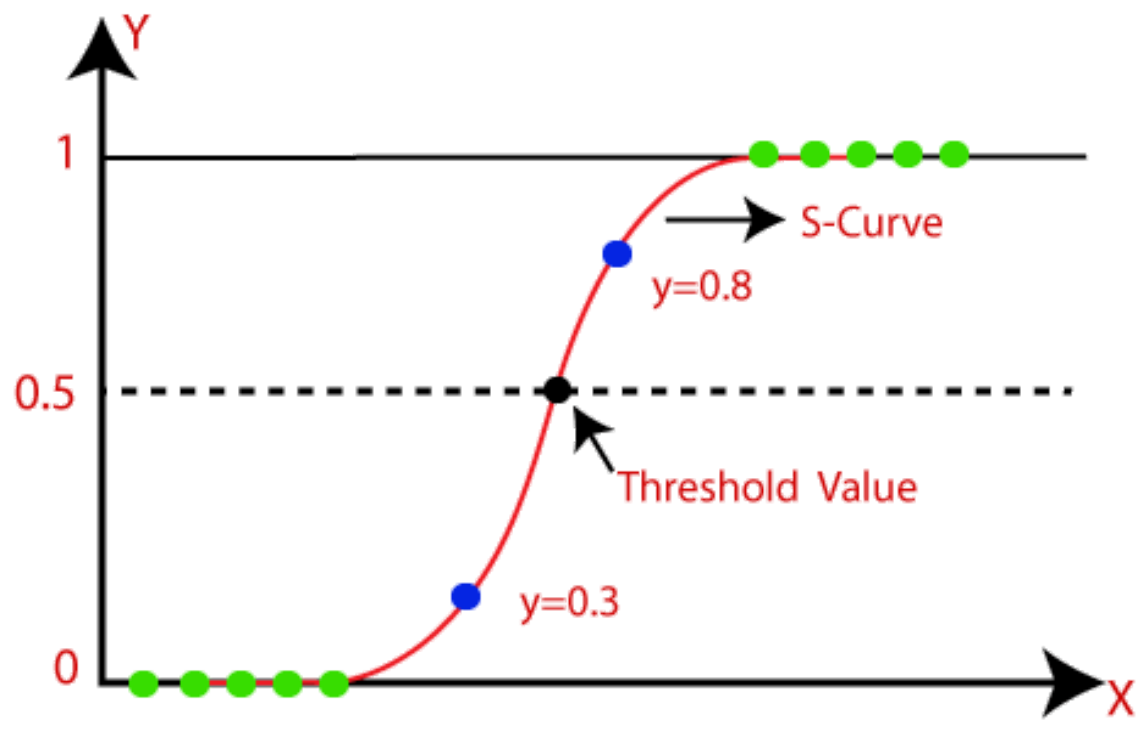

Figure 1. Logistic Regression

## 2.7. Random Forest

Random forest is a machine learning algorithm used intensively for classification and regression. It plants decision trees with varying samples and collects the majority vote for each classification and the result is the average in case of regression [14].

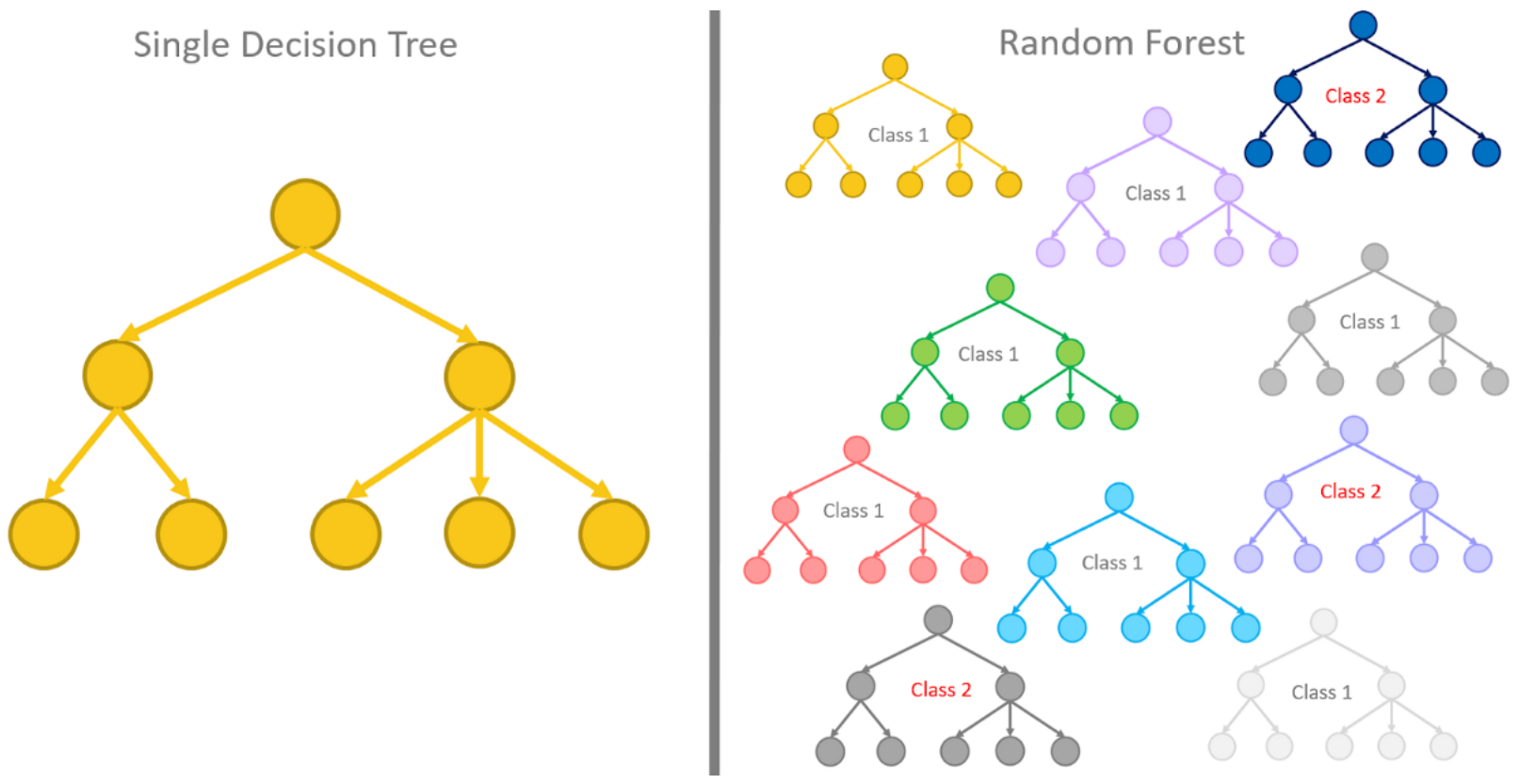

Figure 2. Decision Tree and Random Forest

### 2.8. Multinomial Naive Bayes

The Naive Bayes method of training a model consists of a collection of supervised learning algorithms based on applying the Bayes theorem with the naive hypothesis of conditional independence between each pair of features provided the value of the class variable.

A type of Naive Bayes called Multinomial Naive Bayes is designed well for textual data. Unlike simple naive bayes where the model would just check if a word is present or absent, Multinomial Naive Bayes takes into accord the word count and calculates appropriately. Naive Bayes algorithm is very efficient to implement and hence gives good results for text classification [15].

## 3. Implementation

The purpose of this study is to identify the disaster prone areas using news articles and prevent future disasters. Natural disasters happen more frequently than before due to global warming. Understanding the topography and deficiencies of infrastructure can prevent further disaster. The local government will be able to manage the crisis. Figure 1 shows the flow of implementation.

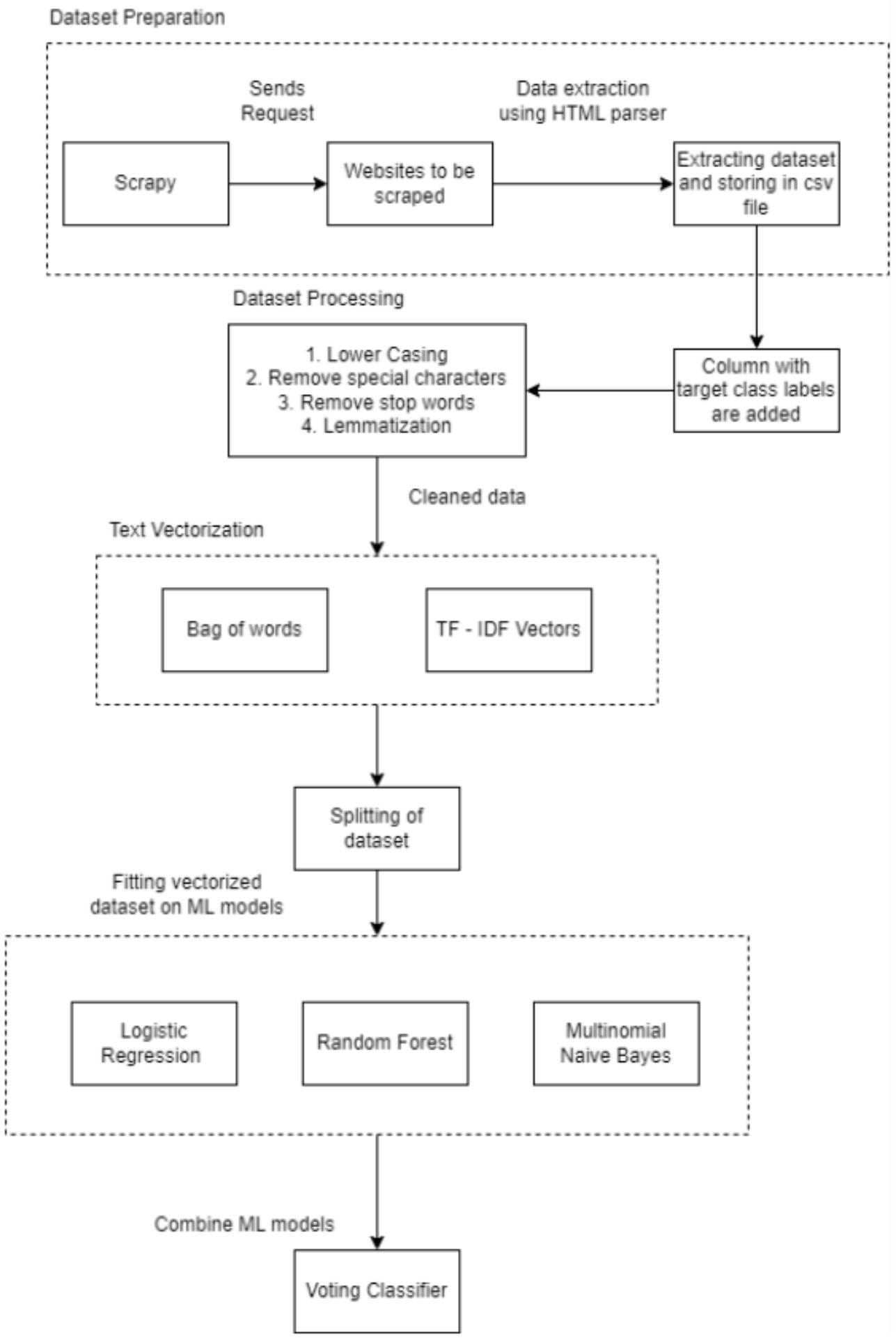

Figure 1. Conceptual Framework

### 3.1. Media Outlets Identification

The News Media Outlets identified for data harvest are the following: SunStart Cebu, Cebu Daily News, Philippines Daily Inquirer for the legitimate news sources and for the illegitimate ones, two media outlets were identified: Adobo Chronicles, a satirical news source with exaggerated articles and dramatic titles. Harvesting of the dataset will be from December 2021 to February 2022. The researchers chose these media outlets to indicate a clear difference between the legitimate and illegitimate news articles.

### 3.2. Data Harvest

Through the use of a web scraping tool called “Scrapy”, the researchers were able to scrape and take relevant information from the news articles such as the title, body of the article, url, date, location, publisher, disaster and author. The tool successfully scraped through the news articles and obtained a total of 160 datasets with 1:1 ratio, 80 from the legitimate news sources such as SunStar Cebu, Cebu Daily News and The Daily Inquirer and 80 for the illegitimate news source Adobo Chronicles.

### 3.3. Machine Learning Model Development

The following machine learning algorithms are implemented; Logistic Regression, Random Forest, Gaussian Naive Bayes, and Multinomial Naive Bayes, to train, test and develop the machine learning model, and these algorithms are combined using the voting classifier technique. Additionally, techniques such as removing the stop words and implementing lemmatization are included in the development of the machine learning model.

### 3.4. Training and Testing

The datasets were split between a 3:7 ratio, 30% of the datasets for the training of the machine learning model and 70% of the datasets for the testing of the machine learning model. The datasets throughout the training and testing phase were balanced with a 1:1 ratio from the legitimate and illegitimate news sources, to keep it consistent between the training phase and testing phase. Additionally, through the training and testing phase, the researchers evaluated the performance through Bag of Words and TF-IDF both lower casing, removing special characters, the stop words and implementing lemmatization.

## 4. Results and analysis

This chapter presents the analysis of the data gathered. Each set of data was analysed under different methodologies and different processes to shed light on the problem and obtain results.

The model utilises the following machine learning algorithms: Logistic Regression, Random Forest Classifier, and Multinomial Naive Bayes in identifying between Legitimate and Illegitimate news articles from the datasets provided. During the training and testing process, the datasets were segregated into 3:7 between the training and testing process stage respectively, each having a 1:1 between legitimate and illegitimate news articles to keep the consistency of the data being fed into the machine learning model. A total of 160 datasets were used between the training and testing phase of the machine learning model. Additionally, a confusion matrix was created. The before mentioned confusion matrix variables were then used to calculate the F-Score (Accuracy, Precision, Recall and F1 Score). Furthermore, the output included the identification of the top 30 frequently used words for the legitimate news articles and illegitimate ones. The results were then used to create a word cloud to showcase the words identified. Table 1 shows the results of all the machine learning algorithms utilised in the machine learning model.

Table 1. Accuracy of the Machine Learning Model(s)

| **Machine Learning Models** | **Accuracy (%)** | |
|---|---|---|
| | **BOW** | **TF-IDF** |
| **Logistic Regression** | 85.71 | 77.68 |
| **Random Forest** | 88.39 | 83.93 |
| **Multinomial Naive Bayes** | 90.18 | 90.18 |
| **Logistic Regression + Random Forest Classifier + Multinomial Naive Bayes** | 91.07 | 90.18 |

The machine learning model displays a high degree of accuracy, even more so for the Combined BOW model which shows a result of 91.07% as well as the Combined TF-IDF model also showed a high degree of accuracy, showing a result of 90.18% even with the balance dataset used in training and testing the machine learning model.

In addition, the model was able to generate the variables needed to identify the F-Score (Precision, Recall, F1-Scored). Table 2 shows results of the calculations of the formulas for the precision, recall and f1-score.

Table 2. F-Score of the Machine Learning Model(s)

| **Machine Learning Models** | **F-Score (%)** | | | | | |
|---|---|---|---|---|---|---|
| | **Precision** | | **Recall** | | **F1-Scored** | |
| | **BOW** | **TF-IDF** | **BOW** | **TF-IDF** | **BOW** | **TF-IDF** |
| **Logistic Regression** | 87.04 | 70.13 | 83.93 | 96.43 | 85.46 | 81.20 |
| **Random Forest** | 81.54 | 81.25 | 94.64 | 92.86 | 87.60 | 86.67 |
| **Multinomial Naive Bayes** | 86.89 | 86.89 | 94.64 | 94.64 | 90.60 | 90.60 |
| **Logistic Regression + Random Forest Classifier + Multinomial Naive Bayes** | 88.33 | 86.89 | 94.64 | 94.64 | 91.38 | 90.60 |

Moreover, a confusion matrix was created to help visualise the performance of the Machine Learning Model by displaying the true label and the predicted label. Figure 2 shows the confusion matrix of the combined machine learning classifiers for BOW and TF-IDF.

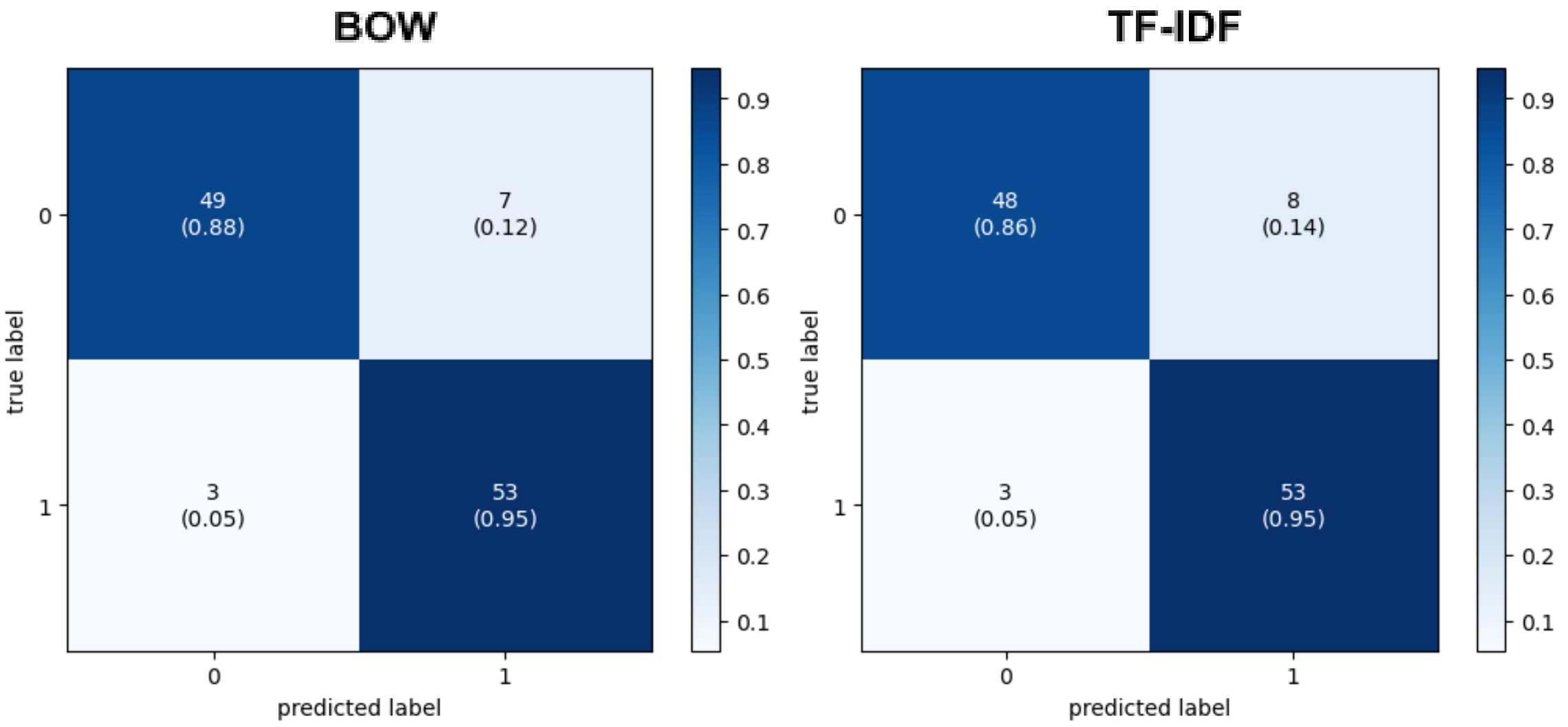

Figure 2. Confusion Matrix for Combined ML Model for BOW and TF-IDF

The confusion matrices show a total of 112 objects from the testing phase. Two confusion matrices were generated from the model, one for the Combined Bag of Words model and one for the Combined TF-IDF Model. The confusion matrices showcase how the data has been classified and show a distinct visual representation of the accuracy of the model for the combined models, 91.07% for the Combined BOW Model and 90.18% for the Combined TF-IDF Model. These confusion matrices indicate that regardless of accuracy, they are reliable in classifying the performance of the model.

Furthermore, with the use of scrapy. The model was able to identify the top 30 frequently used words between the Legitimate News articles and the Illegitimate News articles. These words are detailed and displayed in the figure found below.

| Legitimate news | |
|---|---|
| No. of times | key word |
| 49 | year |
| 48 | website continuing |
| 47 | website |
| 46 | visayas |
| 45 | use cookies |
| 44 | use |
| 43 | town |
| 42 | titles share |
| 41 | titles |
| 40 | subscribe |
| 39 | social media |
| 38 | social |
| 37 | share |
| 36 | said |
| 35 | public |
| 34 | province |
| 33 | president |
| 32 | police |
| 31 | philippines |
| 30 | philippine daily |
| 29 | philippine |
| 28 | office |
| 27 | media |
| 26 | mayor |
| 25 | mandaue |
| 24 | local |
| 23 | listen news |
| 22 | listen |
| 21 | lapu |
| 20 | inquirer |

| Fake news | |
|---|---|
| No. of times | key word |
| 49 | world |
| 48 | vice president |
| 47 | vice |
| 46 | told |
| 45 | support |
| 44 | story |
| 43 | spokesperson |
| 42 | senator |
| 41 | said |
| 40 | sa |
| 39 | robredo |
| 38 | ressa |
| 37 | rappler |
| 36 | rally |
| 35 | presidential aspirant |
| 34 | presidential |
| 33 | president leni |
| 32 | president |
| 31 | pink |
| 30 | philippines manila |
| 29 | philippines |
| 28 | philippine |
| 27 | petition |
| 26 | pacquiao |
| 25 | news |
| 24 | new |
| 23 | na |
| 22 | mayor |
| 21 | marcos |
| 20 | manila bureau |

Figure 3. Top 30 Frequently used Words

Along with this, the researchers were able to generate a word cloud for both the legitimate and illegitimate news articles as a part of the output for the bag of words. Figure 3 and 4, show the word cloud for the legitimate and illegitimate frequently used word respectively.

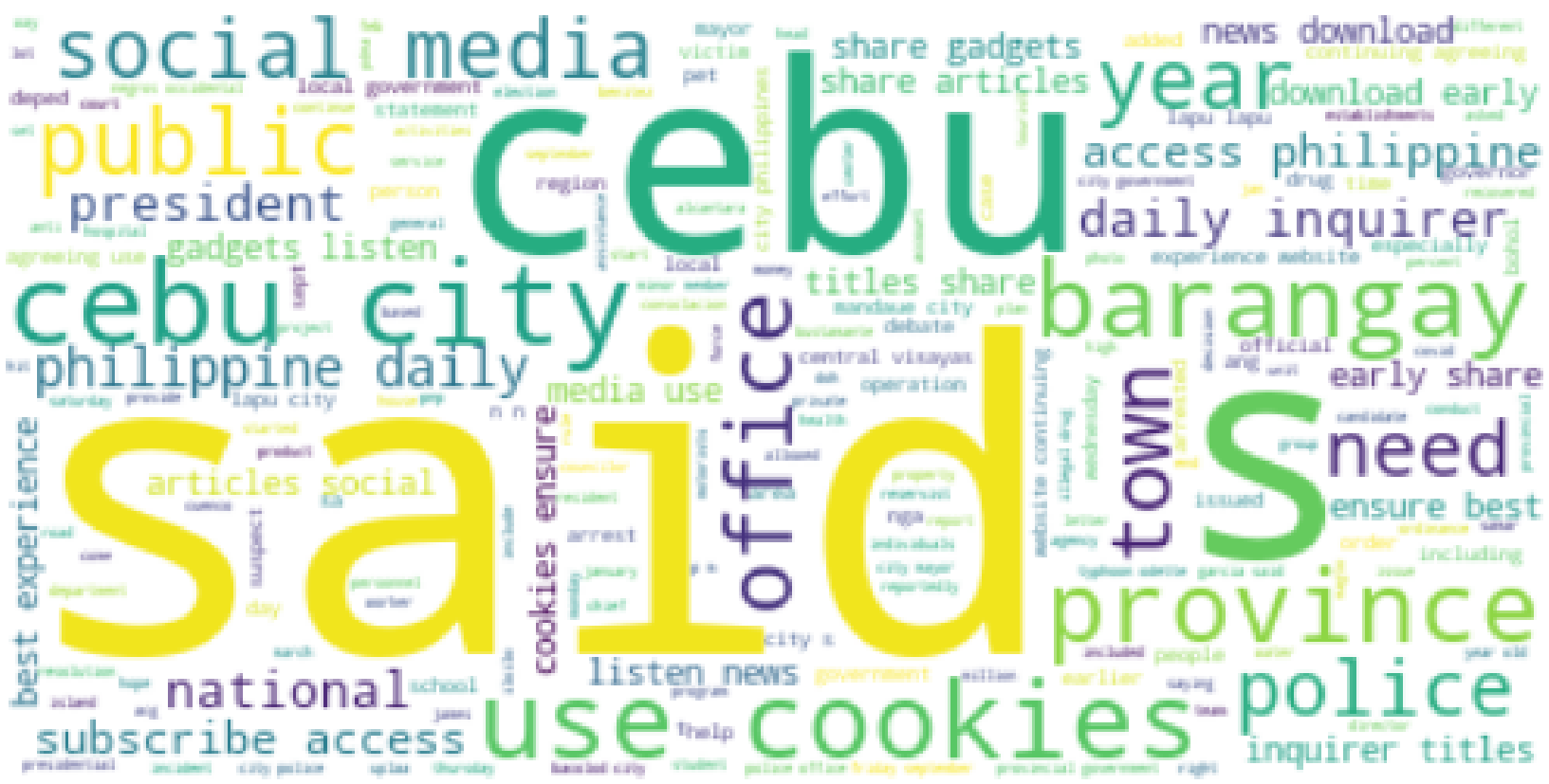

Figure 3. Word Cloud for Legitimate News Articles

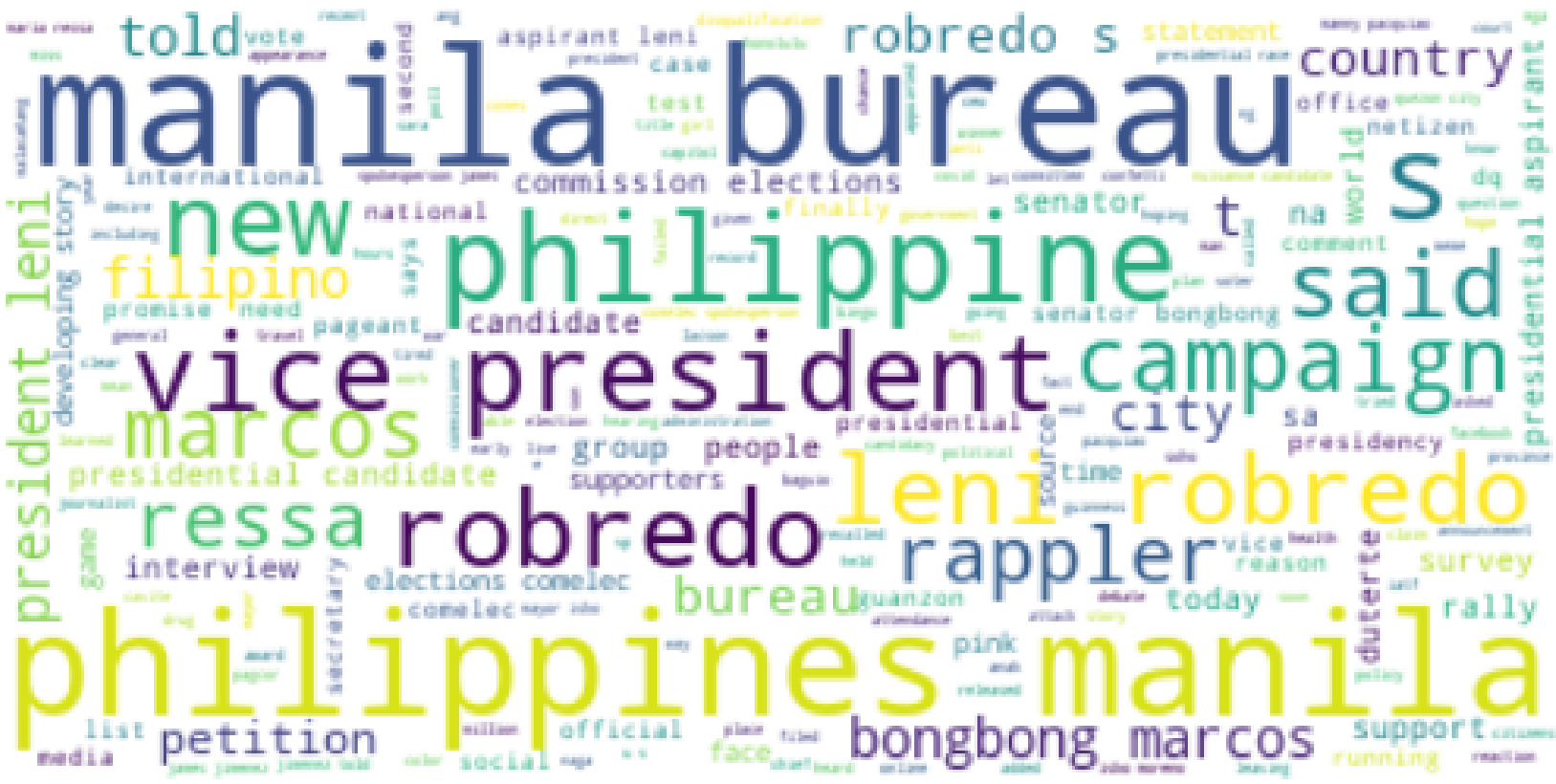

Figure 4. Word Cloud for Illegitimate News Articles

## 4. Conclusion and future scope

In this day and age, the majority of the media we consume is found online. News websites have been an integral part in our day to day lives, they keep us updated on the current events in the country, and especially last December 2021 when Typhoon Odette struck the Philippines leaving the people clamouring for any relevant information on the extent of the disaster. Due to this, opportunistic scam artists capitalise on the situation by publishing fraudulent articles. This created a predicament where the topic of this research was founded on, the identification between legitimate and illegitimate news articles about the disaster last December "Typhoon Odette". In the creation of the Machine Learning Model, the researchers agreed upon using the following algorithms: Logistic Regression, Random Forest Classifier, and Multinomial Naive Bayes. The chosen machine learning algorithms were then trained and tested with the ratio of 3:7 respectively. In addition, a confusion matrix was created identifying the TP (True Positives), TN (True Negatives), FP (False Positives), and FN (False Negatives) and these were then fed through the formulas for identifying the precision, recall and F1 score. The results for the accuracy of 91.07% for the Combined BOW model and 90.18% for the Combined TF-IDF model. F-Scores for the Combined BOW model showed precision of 88.33%, recall of 94.64%, and f1-score of 91.38% and for the Combined TF-IDF model showed precision of 86.89%, recall of 94.64%, and f1-score of 90.60%.

For future studies it is imperative to use additional datasets from multiple news sources, not only centrally focused on the Island of Cebu for training and testing the performance of the model in identifying between legitimate and illegitimate news articles. Additionally, the researchers recommended the use of Passive Aggressive Classifiers used for large-scale learning in the online landscape. It is great in sequentially training the model through batches, which is extremely useful in dealing with high amounts of datasets. Catered specifically for detecting fake news as there are always new articles popping up, updating the model every time it detects a fake news article. Lastly, the researchers recommend semantic analysis. To put it simply, semantic analysis is a natural language processing technique that processes the grammatical structure and relationship between individual words. This is highly valuable in identifying the context between the group of words, extracting meaningful information from unstructured data.

**Authors**

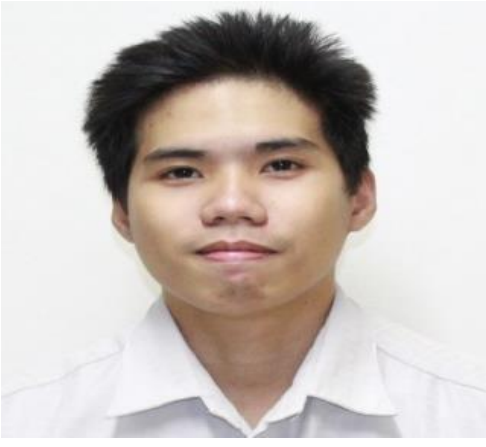

My name is Lord Christian Carl H. Regacho, studying Bachelor of Science in Computer Science at the University of San Carlos..

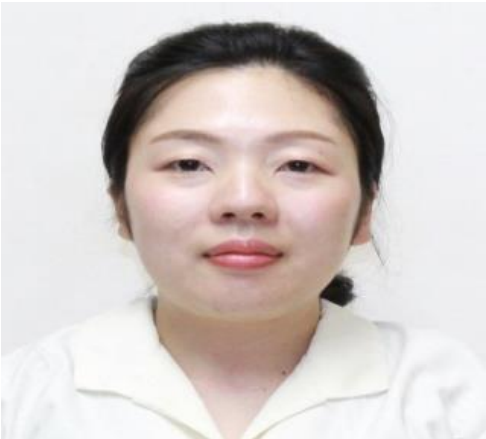

Ms. Ai Matsushita is currently pursuing her Bachelor of Science in Computer Science at the University of San Carlos, Cebu city, the Philippines.

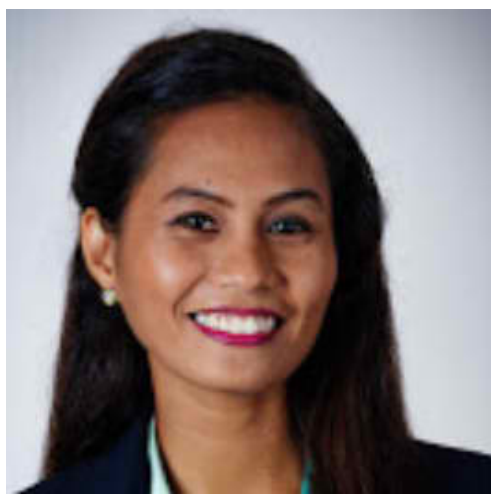

Dr. Angie M. Ceniza-Canillo is a chair of the Department of Computer, Information Sciences and Mathematics at the University of San Carlos, Cebu city, the Philippines.